\documentclass[11pt,letterpaper,logo,twocolumn]{style}

\usepackage[numbers]{natbib}
\usepackage{graphicx}
\usepackage{booktabs}
\usepackage{amsmath,amsfonts,amssymb}
\usepackage{subcaption}
\usepackage{multirow}
\usepackage{colortbl}
\usepackage{listings}
\usepackage{xparse}
\usepackage{fontawesome5}
\usepackage{threeparttable}

\graphicspath{{./}{fig/}{figures/}{plot/}{pdf/}{table/}}
\usepackage{amsthm}
\usepackage{tcolorbox}
\tcbuselibrary{skins,breakable}
\tcbuselibrary{listingsutf8}
\usepackage{titletoc}
\usepackage{pifont}
\usepackage{mathtools}
\usepackage{bbm}
\usepackage{makecell}
\usepackage{adjustbox}

\usepackage{algorithm}
\usepackage{algorithmic}
\usepackage{multirow}
\usepackage{balance}
\usepackage{xcolor}
\usepackage{enumitem}
\usepackage[normalem]{ulem}
\useunder{\uline}{\ul}{}


\title{One Adapts to Any: Meta Reward Modeling for Personalized LLM Alignment}
\runningtitle{One Adapts to Any: Meta Reward Modeling for Personalized LLM Alignment}
\PublicDate{2026-4-20}

\author{%
  {\Authfont
    \textbf{Hongru Cai}\textsuperscript{1} \quad
    \textbf{Yongqi Li}\textsuperscript{1}\advisor \quad
    \textbf{Tiezheng Yu}\textsuperscript{2} \quad
    \textbf{Fengbin Zhu}\textsuperscript{3}\advisor \quad \\
    \Authfont
    \textbf{Wenjie Wang}\textsuperscript{4} \quad
    \textbf{Fuli Feng}\textsuperscript{4} \quad
    \textbf{Wenjie Li}\textsuperscript{1} \quad
  }\\
  {\Affilfont
    \textsuperscript{1} The Hong Kong Polytechnic University \quad
    \textsuperscript{2} Huawei Technologies Ltd. \quad \\
     \textsuperscript{3} National University of Singapore \quad
    \textsuperscript{4} University of Science and Technology of China \quad \\
    \texttt{\{henry.hongrucai, liyongqi0, zhfengbin\}@gmail.com}
  }
}

\begin{document}

\begin{abstract}
Alignment of Large Language Models (LLMs) aims to align outputs with human preferences, and personalized alignment further adapts models to individual users. This relies on personalized reward models that capture user-specific preferences and automatically provide individualized feedback. However, developing these models faces two critical challenges: the scarcity of feedback from individual users and the need for efficient adaptation to unseen users. We argue that addressing these constraints requires a paradigm shift from fitting data to learn user preferences to learn the process of preference adaptation. To realize this, we propose \textbf{M}eta \textbf{R}eward \textbf{M}odeling (MRM), which reformulates personalized reward modeling as a meta-learning problem. Specifically, we represent each user's reward model as a weighted combination of base reward functions, and optimize the initialization of these weights using a Model-Agnostic Meta-Learning (MAML)-style framework to support fast adaptation under limited feedback. To ensure robustness, we introduce the \textbf{R}obust \textbf{P}ersonalization \textbf{O}bjective (RPO), which places greater emphasis on hard-to-learn users during meta optimization. Extensive experiments on personalized preference datasets validate that MRM enhances few-shot personalization, improves user robustness, and consistently outperforms baselines.
\end{abstract}

\newcommand{\TitleLinks}{%
\centering
    \vspace{6pt}
    {\noindent\absfont\fontsize{11}{13}\selectfont
    \faGithub\ Project Page: \url{https://github.com/ModalityDance/MRM}\par}%
}


\maketitle

\section{Introduction}

\begin{figure}[t]
\setlength{\abovecaptionskip}{-0.05cm}
\setlength{\belowcaptionskip}{-0.2cm}
\centering
\includegraphics[width=1\linewidth]{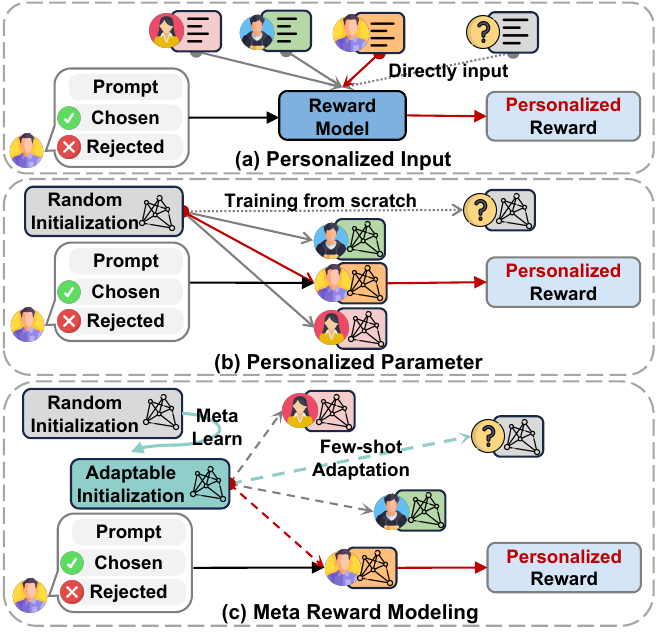}
\caption{Comparison of personalized reward modeling methods: (a) Personalized input incorporates user contexts; (b) Personalized parameter assigns user-specific parameters; (c) Meta Reward Modeling formulates personalization as a meta-learning problem by learning an adaptable initialization.}
\label{Fig: intro}
\end{figure}

The goal of Large Language Model (LLM) alignment is to ensure that models behave consistently with human preferences~\cite{wang2023aligninglargelanguagemodels}. Previous approaches typically assume a uniform standard and optimize for generic preferences that are broadly acceptable~\cite{kaufmann2025surveyofrlhf}. However, human values are highly diverse. Preferences vary significantly across individuals, which makes a single standard insufficient~\cite{guan-etal-2025-surveypersonalizedalignment}. Therefore, recent work emphasizes personalized alignment to address the plurality of user intents and contexts~\cite{araodmap}. By tailoring model behavior to individual needs, personalized alignment enables LLMs to respect unique user preferences rather than following a single monolithic objective~\cite{zhang2025personalizationLLMsurvey, guan-etal-2025-surveypersonalizedalignment}.

The alignment of LLMs relies on human feedback to provide learning signals that guide model behaviors. Standard approaches, such as Reinforcement Learning from Human Feedback (RLHF)~\cite{2017RLHF}, optimize LLMs by comparing responses ranked by human preference~\cite{2017RLHF, bai2022traininghelpfulharmlessassistant, ouyang2022training,li2025harmlessmultimodalassistantsblind}. However, keeping a human-in-the-loop to provide continuous feedback is impractical~\cite{kaufmann2025surveyofrlhf, rafailov2023dpo}. This challenge is even greater in personalized alignment, as we cannot expect every single user to actively provide feedback to align the model. This drives the research on personalized reward models, which learn individual preferences and provide personalized feedback automatically~\cite{poddar2024VPL,chen2025PAL, ryan-etal-2025-synthesizeme}.

While personalized reward models are essential for achieving personalized alignment, their development still faces unique challenges. 1) \textbf{Scarcity of personalized feedback}. General reward models typically rely on massive preference data to capture preference patterns~\cite{liu2025skyworkrewardv2}, but gathering such extensive feedback from a single user is impractical. Simply aggregating data from other users is ineffective, as their conflicting preferences would confuse the model. Thus, the personalized feedback available for each user is inherently sparse, making it challenging to model their unique intent. 2) \textbf{Adaptation to unseen users}. It is impractical to collect data from every potential user in advance. As a result, personalized reward models must handle unseen users\footnote{In this work, ``unseen users'' refer to users who are not included in the training phase but provide a few preference examples during the test phase for personalization.} whose feedback was not pre-collected, and quickly adapt to their unique preferences using only a few demonstrations~\cite{bose2025LORE}. Both the scarcity of individual feedback and the necessity of adapting to unseen users pose significant challenges to personalized reward modeling.

Existing approaches to personalized reward modeling typically rely on either personalized inputs or personalized parameters, but both face limitations. 1) \textbf{\textit{Personalized input}} methods (see Figure~\ref{Fig: intro} (a)) achieve personalization by explicitly incorporating user context into one shared model, such as persona descriptions~\cite{zhang-2024GPG, ryan-etal-2025-synthesizeme}, prior preferences~\cite{zhao2024GPO}, or learned embeddings~\cite{poddar2024VPL}. However, since the model parameters are fixed, personalization relies entirely on the quality of the input context. This fails under sparse feedback, as the limited context cannot distinguish unique user intents, forcing the model into a coarse-grained approximation~\cite{zhao2024GPO} instead of fine-grained personalization. 2) \textbf{\textit{Personalized parameter}} methods (see Figure~\ref{Fig: intro} (b)) allocate user-specific parameters, such as LoRA adapters or separate models~\cite{chen2025PAL, rame2023rewardedsoups, jang2023personalizedsoupspersonalizedlarge}. Although this allows for fine-grained alignment, it struggles to adapt to unseen users. Training user-specific parameters from scratch is infeasible in the few-shot scenario, as sparse feedback inevitably leads to overfitting and poor generalization. Consequently, neither paradigm can simultaneously efficiently adapt to unseen users while maintaining fine-grained personalization under limited user feedback.

This dilemma implies that current optimization paradigms are inadequate and calls for a fundamental shift in perspective. In this work, we reframe personalized reward modeling through the lens of meta-learning, treating each individual's preference modeling as a distinct learning task. Instead of relying on fitting data to learn each individual preference, we expect that the model could learn the process of preference adaptation (i.e., ``learning to learn''). Specifically, we implement this idea by learning a highly adaptable model initialization from users (see Figure~\ref{Fig: intro} (c)). We employ a bi-level optimization framework: the inner loop mimics the adaptation to individual users using sparse feedback, while the outer loop updates the initialization to achieve faster adaptation across users. By learning from this adaptation process, the model captures intrinsic preference commonalities across diverse users and thus serves as a starting point that can rapidly converge to any user's intent.

Building on this insight, we propose \textbf{M}eta \textbf{R}eward \textbf{M}odeling (MRM). To enable lightweight per-user adaptation, we model each user reward as a low-dimensional weight combination over shared basis reward functions~\cite{shenfeld2025PREF,bose2025LORE} and employ a Model-Agnostic Meta-Learning (MAML)~\cite{2017maml}-style framework to optimize the initialization of the weights. However, given the diversity of human values, some users are naturally harder to model than others. Standard meta-learning treats all users equally, aiming to maximize the average performance. Consequently, this uniform approach often neglects users with unique or complex needs, as the model prioritizes the majority to reduce the overall error. To address this, we introduce the \textbf{R}obust \textbf{P}ersonalization \textbf{O}bjective (RPO). RPO dynamically gives more weight to these hard-to-learn users during training, identified by their meta-training losses. This prevents the model from ignoring distinct preferences, ensuring consistent and robust performance for all users. Together, our framework enables learning from sparse feedback and efficient adaptation to unseen users, while maintaining robustness across diverse preferences.

The key contributions of this work are as follows:
\begin{itemize}[itemsep=0pt, topsep=0pt, parsep=0pt, partopsep=0pt, leftmargin=*]
\item We introduce a new formulation of personalized reward modeling from a meta-learning perspective, where each user is treated as a distinct task. This formulation explicitly targets rapid adaptation from sparse feedback and generalization to unseen users, addressing fundamental limitations of existing paradigms.

\item We propose Meta Reward Modeling, which learns a shared and highly adaptable initialization over a set of base reward functions for efficient few-shot personalization. In addition, we design a Robust Personalization Objective that emphasizes hard-to-learn users during meta-optimization, improving robustness under diverse user preferences.

\item We demonstrate through extensive experiments that MRM consistently outperforms baselines in adaptation and robustness, thereby advancing the frontier of personalized LLM alignment.
\end{itemize}
\section{Related Work}

In this section, we revisit prior studies on reward models, personalized LLMs, and meta-learning for personalization. 

\vspace{2pt}
\noindent$\bullet$ \textbf{Reward Models.}
Reward models are a key component in aligning LLMs, as they translate human preferences into feedback signals that guide optimization~\cite{zhong2025comprehensivesurveyrewardmodels}. Their design can be characterized along two dimensions. 1) At the \textbf{type level}, there are three categories. Discriminative models~\cite{cai2024internlm2technicalreport, yuan2025advancing, liu2024skyworkrewardbagtricksreward, wang-etal-2024-interpretable} output a scalar score through a prediction head. Generative models~\cite{zheng2023llmasajudge, li2024generative, cao2024compassjudger1allinonejudgemodel, ye2025learningLLMasajudge} produce natural language judgments before mapping them into a score. Implicit models~\cite {zhao2023slichfsequencelikelihoodcalibration, rafailov2024from,gheshlaghi-azar24ipo, kto} define rewards through generation probabilities to directly optimize on preference pairs. 2) At the \textbf{granularity} level, outcome-based models~\cite{zhu2024starlingb, liu2024skyworkrewardbagtricksreward, yang2024regularizing} assign a score to the entire response, while process-based models~\cite{li-etal-2023-making, uesato2023solving, lightman2024lets, wang-etal-2024-math, wang-etal-2024-multi-step} provide supervision at each intermediate step. However, these approaches generally assume a single standard of human preference, overlooking the personalized preferences of users.

Personalized reward models aim to capture individual differences in user preferences rather than assuming a single universal standard~\cite{araodmap, guan-etal-2025-surveypersonalizedalignment}. Existing methods can be broadly divided into two categories. 1) \textbf{Personalized input} approaches condition the model on user-specific signals derived from user history, such as persona descriptions~\cite{zhang-2024GPG, ryan-etal-2025-synthesizeme}, prior preferences~\cite{zhao2024GPO}, or learned embeddings~\cite{poddar2024VPL}. While these approaches avoid training separate models, they rely entirely on the quality of the input context. Under sparse feedback, they struggle to capture unique user intents, often resulting in coarse-grained approximations~\cite{zhao2024GPO} rather than fine-grained personalization. 2) \textbf{Personalized parameter} approaches assign user-specific modules or parameters to capture individual preferences~\cite{chen2025PAL,rame2023rewardedsoups, jang2023personalizedsoupspersonalizedlarge, bose2025LORE,shenfeld2025PREF}. Although allowing for more flexible alignment, training these user-specific parameters from scratch is infeasible in few-shot scenarios. This paradigm is prone to overfitting when feedback is scarce and fails to generalize efficiently to unseen users. In contrast to these static fitting paradigms, our approach reformulates personalized reward modeling through the lens of meta-learning. By learning highly adaptable initialization and shared preference patterns, our method enables rapid and robust adaptation to unseen users using only minimal feedback.

\vspace{2pt}
\noindent$\bullet$ \textbf{Personalized LLMs.}
Personalized LLMs are designed to meet the individualized needs of distinct users~\cite{tseng-etal-2024-two}. Research can be categorized into two main areas. 1) \textbf{Personalized content generation} focuses on generating personalized content. They have used openly available user data on Reddit~\cite{welch-etal-2022-leveraging}, Twitter~\cite{shokri2017membership}, and other blogging websites~\cite{king-cook-2020-evaluating} to pre-train LLMs. Key tasks include stance classification, demographic inference~\cite{soni-etal-2022-human}, and personalized sentiment prediction~\cite{mireshghallah-etal-2022-useridentifier}. 2) \textbf{Applications in real-world scenarios} starting with personalized dialogue systems~\cite{2025personalwab, you2025r2eclargerecommendermodels, xiao2026alpsbenchllmpersonalizationbenchmark}. Studies have built datasets based on specific personas~\cite{zhang-etal-2018-personalizing}, and by extracting user attributes from Reddit~\cite{mazare-etal-2018-training} and Weibo~\cite{zhong-etal-2022-less}. Applications also include fields like healthcare~\cite{yu2025healthllmpersonalizedretrievalaugmenteddisease}, education~\cite{shehata-etal-2023-enhancing}, and robotics ~\cite{personalizedrobot}. While these studies mainly operate at the policy or generation level, our work instead focuses on personalized reward modeling, providing an alternative path that enables few-shot personalization.

\vspace{2pt}
\noindent$\bullet$ \textbf{Meta-Learning for Personalization.}
Meta-learning focuses on training models that can rapidly adapt to new tasks with only a few samples. There are three categories: model-based methods~\cite{modelbasemetalearning} that design architectures with fast adaptation capabilities, metric-based methods~\cite{metricbased} that learn similarity measures for generalization, and optimization-based methods~\cite{2017maml} that learn effective initialization. Our work is based on the third paradigm of MAML~\cite{2017maml}, which learn good initializations for rapid adaptation. Meta-learning has been applied in many domains including image classification~\cite{modelbasemetalearning, 2017maml, mishra2018a}, language modeling~\cite{chen-etal-2022-meta, min-etal-2022-metaicl}, and reinforcement learning~\cite{duan2016rl2fastreinforcementlearning, wang2017learningreinforcementlearn}. 

In personalization, users are often treated as tasks within the meta-learning framework~\cite{madotto-etal-2019-personalizingdialogue, 2024limamal}. Works in recommendation~\cite{2017metalearncoldrecitem,2024limamal,2023m2eu, 2023metaadaptiveloss, 2019melu} showed that meta-learning can help to adapt to new users and thus alleviate cold-start problems. Similar ideas have been applied to LLMs~\cite{zollo2025personalllm, zhao2025metalearningcoldstartpersonalizationprompttuned}to help adapt models to diverse preferences. FSPO~\cite{singh2025fspo} combines meta-learning with DPO~\cite{rafailov2023dpo} to learn personalized preferences at the policy level, directly optimizing the model’s generation behavior. In contrast, our work focuses on personalized reward modeling, learning user-specific reward functions that capture diverse user preferences.
\section{Preliminaries}

Before presenting our proposed method, we first review two key components that help to understand our method: Model-Agnostic Meta-Learning~\cite{2017maml}, and the training of reward models.

\subsection{Model-Agnostic Meta-Learning}

The core idea of MAML is to learn a set of initial parameters that can be efficiently adapted to new tasks using only a few examples. As shown in Algorithm~\ref{alg:maml}, let the model be $f_\theta$ with parameters $\theta$. Each task $\mathcal{T}_i$ is drawn from a task distribution $p(\mathcal{T})$ and consists of two parts: a support set $\mathcal{D}_i^s$ used for task-specific adaptation, and a query set $\mathcal{D}_i^q$ used for evaluation and optimization.

\vspace{3pt}
\noindent$\bullet$ \textbf{Inner loop task adaptation.} In this phase, the model adapts to task $\mathcal{T}_i$ by updating $\theta$ with a few gradient steps on the support set:

\begin{equation}
\theta_i = \theta - \alpha \nabla_\theta \mathcal{L}_{\mathcal{T}_i}\big(f_\theta, \mathcal{D}_i^s\big),
\end{equation}
where $\alpha$ is the inner learning rate, and $\mathcal{L}_{\mathcal{T}_i}$ is the task-specific loss (e.g., cross-entropy~\cite{baik2021metataskadaptiveloss} or MSE~\cite{metalearnedloss}). After this step, $\theta_i$ represents the personalized parameters for task $\mathcal{T}_i$, adapted from a shared initialization.

\vspace{3pt}
\noindent$\bullet$ \textbf{Outer loop meta optimization.} In this phase, the adapted parameters $\theta_i$ are evaluated on the query set, and the shared initialization $\theta$ is updated to improve generalization across tasks:

\begin{equation}
\theta \leftarrow \theta - \beta \nabla_\theta \sum_i \mathcal{L}_{\mathcal{T}_i}\big(f_{\theta_i}, \mathcal{D}_i^q\big),
\end{equation}
where $\beta$ is the meta learning rate. This update requires differentiating through the inner loop with respect to the initialization parameters.

Through this two-level optimization, MAML learns an initialization $\theta$ that can be rapidly personalized to new tasks with only a few examples. This ability to support fast adaptation and few-shot learning makes MAML especially well-suited for personalized reward modeling, where limited feedback is available for each user.

\begingroup
\setlength{\textfloatsep}{0pt}
\setlength{\intextsep}{0pt}
\setlength{\floatsep}{0pt}
\begin{algorithm}[t]
\caption{Model-Agnostic Meta-Learning~\cite{2017maml}}
\label{alg:maml}
\footnotesize
\begin{algorithmic}[1]
\REQUIRE $p(\mathcal{T})$: distribution over tasks
\REQUIRE $\alpha, \beta$: step sizes for inner and outer updates
\REQUIRE $n$: number of inner loop gradient updates
\STATE Randomly initialize $\theta$
\WHILE{not done}
    \STATE Sample batch of tasks $\mathcal{T}_i \sim p(\mathcal{T})$
    \FORALL{$\mathcal{T}_i$}
        \STATE \textbf{\textcolor{blue}{Inner loop task adaptation:}}
        \STATE $\theta_i \gets \theta$
        \FORALL{ \textbf{$n$ steps} } 
        \STATE Evaluate $\nabla_{\theta_i} \mathcal{L}_{\mathcal{T}_i}\big(f_\theta, \mathcal{D}_i^s\big)$  \hfill \textcolor{blue}{$\triangleright~\mathcal{L}_{\mathcal{T}_i}$: Task-sepcific loss}
        \STATE $\theta_i \gets \theta_i - \alpha \nabla_{\theta_i} \mathcal{L}_{\mathcal{T}_i}\big(f_\theta, \mathcal{D}_i^s\big)$  
        \ENDFOR
    \ENDFOR
    \STATE \textbf{\textcolor{blue}{Outer loop meta optimization:}}
    \STATE Update $\theta \gets \theta - \beta \nabla_\theta \sum_{\mathcal{T}_i} \mathcal{L}_{\mathcal{T}_i}\big(f_\theta, \mathcal{D}_i^q\big)$ 
\ENDWHILE
\end{algorithmic}
\end{algorithm}
\endgroup

\subsection{Reward Model Training}
\label{sec: rm training}
A central step in RLHF~\cite{2017RLHF} is learning a reward function that encodes human preferences. The reward model is trained from pairwise comparisons: for the same prompt, annotators indicate which response they prefer. Formally, the reward model $r_\phi(x, y)$ assigns a scalar score to response $y$ given prompt $x$. For a preference pair $(x, y^+, y^-)$, where $y^+$ is preferred over $y^-$, the Bradley–Terry model~\cite{bradley1952BTloss} defines:

\begin{equation}
P(y^+ \succ y^- \mid x) = \sigma\!\big(r_\phi(x, y^+) - r_\phi(x, y^-)\big),
\end{equation}
with $\sigma(z) = \tfrac{1}{1+\exp(-z)}$. The training loss function is:

\begin{equation}
\mathcal{L}_{\text{RM}}(\phi) = - \sum_{(x, y^+, y^-) \in \mathcal{D}} \log \sigma\!\big(r_\phi(x, y^+) - r_\phi(x, y^-)\big).
\end{equation}
This objective assigns higher scores to preferred responses and lower scores to rejected ones. In personalized scenarios, each user provides only a few comparisons. Training a separate model per user causes overfitting, while a single shared model fails to capture individual preferences. We thus frame personalized reward learning as a meta-learning problem that learns a shared initialization while still enabling user-specific adaptation.

\section{Method}

We introduce our proposed \textbf{Meta Reward Modeling}, which consists of two key components:

\noindent$\bullet$ \textbf{Meta Weight Initialization.} We formulate personalized reward modeling as a meta-learning problem by learning a shared initialization across users. This initialization enables fast personalization from scarce feedback and generalization to unseen users.

\noindent$\bullet$ \textbf{Robust Personalization Objective.} To improve robustness across user diversity, we introduce the Robust Personalization Objective, which reweights user-level query losses to emphasize hard-to-learn users in outer loop meta optimization.

\subsection{Meta Weight Initialization}

\begin{figure}[t]
\setlength{\abovecaptionskip}{-0.05cm}
\setlength{\belowcaptionskip}{-0.20cm}
\centering
\includegraphics[width=1\linewidth]{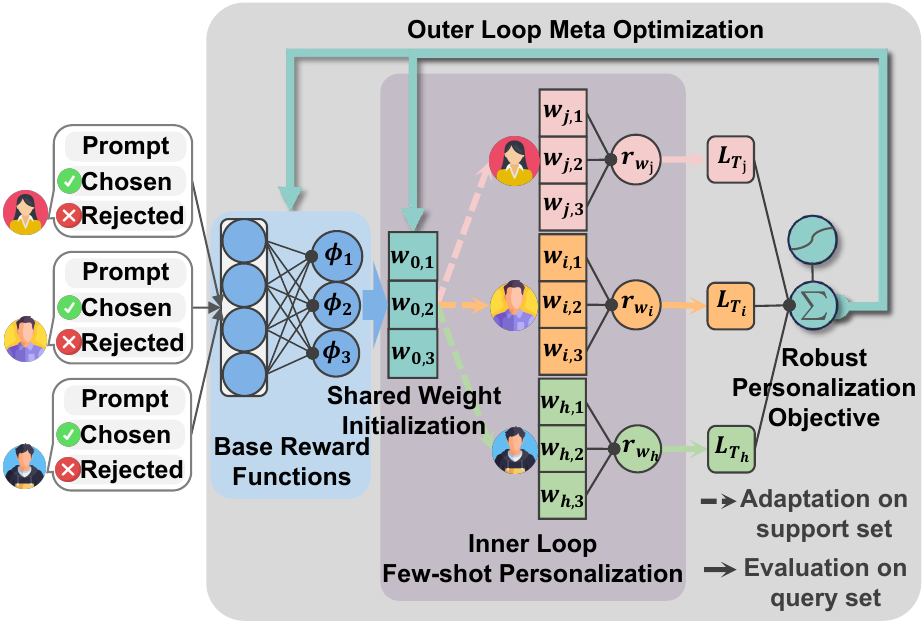}
\caption{Overview of Meta Reward Modeling. The model employs base reward functions with shared weight initialization, adapts user-specific weights in the inner loop, and updates both initialization and base functions in the outer loop with the robust personalization objective.}
\label{Fig: method}
\end{figure}

As shown in Figure~\ref{Fig: method}, the key idea is to represent the personalized reward model as a weighted combination of base reward functions, and to meta-learn a shared initialization of these weights, while keeping the base functions shared across users.

\vspace{3pt}
\noindent$\bullet$ \textbf{Model structure.}
To capture diverse user preferences, following prior work~\cite{bose2025LORE}, we represent the reward model $r_{w_i}$ as a weighted combination of multiple base reward functions. Let ${\phi_k}_{k=1}^K$ denote the set of base reward functions. For a given user $i$, the personalized reward is defined as:

\begin{equation}
r_{w_i}(x,y) = \sum_{k=1}^K w_{i,k}\, \cdot \phi_k(x,y),
\end{equation}
where $w_i = (w_{i,1}, \ldots, w_{i,K})$ are the user-specific weights. The weights $w_i$ determine the contribution of each base function to the overall personalized reward for user $i$. 

\vspace{3pt}
\noindent$\bullet$ \textbf{Task definition.}
Learning the user-specific weights is the core of personalized reward modeling. We treat each user’s weight learning as an independent meta-learning task. Let $\mathcal{T}_i$ denote the task for user $i$, with dataset $\mathcal{D}_i$ containing pairwise preferences in the form of triples $(x, y^+, y^-)$, where $x$ is a prompt, $y^+$ the preferred response, and $y^-$ the less preferred one. Following the meta-learning paradigm, we split $\mathcal{D}_i$ into two disjoint subsets. The support set $\mathcal{D}_i^s$ provides limited feedback for inner-loop adaptation, while the query set $\mathcal{D}_i^q$ serves as held-out data to evaluate the adapted model and update the shared initialization in the outer loop. This separation prevents the model from overfitting to the limited feedback within each user and enables it to learn a shared initialization that supports few-shot adaptation across users.

\vspace{3pt}
\noindent$\bullet$ \textbf{Task loss.}
The training loss for each task is defined by the reward modeling loss introduced in \S~\ref{sec: rm training}. For a preference pair $(x, y^+, y^-) \in \mathcal{D}_i$, the reward function $r_{w_i}(x,y)$ assigns a score to a response given the prompt, and the Bradley–Terry~\cite{bradley1952BTloss} model defines the preference probability as:

\begin{equation}
P(y^+ \succ y^- \mid x, r_{w_i}) = \sigma\big(r_{w_i}(x,y^+) - r_{w_i}(x,y^-)\big),
\end{equation}
where $\sigma(z)$ is the sigmoid function. The loss for user $i$ is then:

\begin{equation}
\mathcal{L}_{\mathcal{T}_i}(w_i) = - \sum_{(x,y^+,y^-)\in \mathcal{D}_i} 
\log \sigma\big(r_{w_i}(x,y^+) - r_{w_i}(x,y^-)\big).
\end{equation}
When restricted to the support set, this loss guides the adaptation of the weights for user $i$. When evaluated on the query set, it provides the signal to improve the shared initialization across users. 

\begingroup
\setlength{\textfloatsep}{0pt}
\setlength{\intextsep}{0pt}
\setlength{\floatsep}{0pt}
\begin{algorithm}[t]
\caption{Meta Reward Modeling}
\label{alg:mrm}
\footnotesize
\begin{algorithmic}[1]
\REQUIRE $p(\mathcal{T})$: distribution over users (tasks)
\REQUIRE $\alpha, \beta$: step sizes for inner and outer updates
\REQUIRE $n$: number of inner loop gradient updates
\STATE Randomly initialize weights $w_0$ and base functions $\{\phi_k\}_{k=1}^K$
\WHILE{not done}
    \STATE Sample batch of user tasks $\mathcal{T}_i \sim p(\mathcal{T})$
    \FORALL{$\mathcal{T}_i$}
        \STATE \textbf{\textcolor{blue}{Inner loop few-shot personalization:}}
        \STATE $w_i \gets w_0$
        \FORALL{ \textbf{$n$ steps} } 
        \STATE Evaluate $\nabla_{w_i} \mathcal{L}_{\mathcal{T}_i}(w_i,\mathcal{D}_i^s)$ \hfill \textcolor{blue}{$\triangleright~\mathcal{L}$: Bradley-Terry Loss~\cite{bradley1952BTloss}}
        \STATE $w_i \gets w_i - \alpha \nabla_{w_i} \mathcal{L}_{\mathcal{T}_i}(w_i,\mathcal{D}_i^s)$ 
        \ENDFOR
    \ENDFOR
    \STATE \textbf{\textcolor{blue}{Outer loop meta optimization:}} 
    \STATE $w_0 \gets w_0 - \beta \nabla_{w_0} \,\mathcal{A}\big(\{\mathcal{L}_{\mathcal{T}_i}(w_i,\mathcal{D}_i^q)\}_i\big)$
    \STATE \parbox[t]{\dimexpr\linewidth-\algorithmicindent}{%
        $\phi_k \gets \phi_k - \beta \nabla_{\phi_k} \,\mathcal{A}\big(\{\mathcal{L}_{\mathcal{T}_i}(w_i,\mathcal{D}_i^q)\}_i\big), \quad k=1,\ldots,K$\\
        \makebox[\linewidth][r]{\textcolor{blue}{$\triangleright$ $\mathcal{A}(\cdot)$: Robust Personalization Objective}}%
    }
\ENDWHILE
\end{algorithmic}
\end{algorithm}
\endgroup

\vspace{2pt}
\noindent$\bullet$ \textbf{Inner loop few-shot personalization.}
As shown in Algorithm~\ref{alg:mrm}, the inner loop starts from the shared initialization $w_0$ and adapts only the user-specific weights $w_i$ for each task $\mathcal{T}_i$ using its support set $\mathcal{D}_i^s$. This step personalizes the reward model for user $i$ with the limited feedback available. The procedure is written as:

\begin{equation}
w_i = w_0, \quad \quad w_i = w_i - \alpha \nabla_{w_i} \mathcal{L}_{\mathcal{T}_i}(w_i,\mathcal{D}_i^s),
\end{equation}
where $\alpha$ is the inner learning rate, and $\mathcal{L}_{\mathcal{T}_i}$ is the task loss defined earlier. In practice, this update can be repeated for a few steps. After adaptation, $w_i$ represents the personalized weights that define the personalized reward function $r_{w_i}$ for this user.

\vspace{2pt}
\noindent$\bullet$ \textbf{Outer loop meta optimization.}
After adapting user-specific weights $w_i$ in the inner loop, we evaluate the personalized weights on the query sets $\mathcal{D}_i^q$. The gradients from these evaluations are then used to update both the shared initialization of the weights and the shared base reward functions. The initialization $w_0$ is updated to provide a better starting point for adaptation:
\begin{equation}
w_0 = w_0 - \beta \nabla_{w_0} \, \mathcal{A}\big(\{\mathcal{L}_{\mathcal{T}_i}(w_i,\mathcal{D}_i^q)\}_{i}\big),
\end{equation}
where $\beta$ is the meta learning rate and $\mathcal{A}(\cdot)$ denotes the robust personalization objective that reweights user-level query losses (see \S\ref{sec:robust_objective}). Similarly, each base reward function is updated:
\begin{equation}
\phi_k = \phi_k - \beta \nabla_{\phi_k} \, \mathcal{A}\big(\{\mathcal{L}_{\mathcal{T}_i}(w_i,\mathcal{D}_i^q)\}_{i}\big), \quad k=1,\ldots,K.
\end{equation}
Thus, the reweighted query losses jointly update $w_0$ and ${\phi_k}_{k=1}^K$, producing initializations that are both adaptable and robust.

\vspace{3pt}
\noindent$\bullet$ \textbf{MRM inference.}
At inference time, the learned initialization $w_0$ enables efficient few-shot personalization. Given a user $u$ with training data $\mathcal{D}_u$, the user-specific weights are adapted from $w_0$ by performing one or a few gradient steps on $\mathcal{D}_u$:
\begin{equation}
w_u = w_0 - \alpha \nabla_{w_0} \mathcal{L}_{\mathcal{T}_u}(w_0, \mathcal{D}_u).
\end{equation}
This produces the personalized weights $w_u$ that define the user-specific reward function:
\begin{equation}
r_{w_u}(x,y) = \sum_{k=1}^K w_{u,k}, \phi_k(x,y).
\end{equation}
Once adapted, $r_{w_u}$ can be directly used to evaluate or rank responses according to the user’s preferences. Benefiting from the meta-learned initialization, MRM achieves both few-shot adaptation within users and strong generalization across users.

\subsection{Robust Personalization Objective}
\label{sec:robust_objective}

While meta weight initialization enables fast personalization from limited data, it overlooks robustness across users. Some users are easy to adapt, while hard-to-learn users with unique or inconsistent feedback perform much worse when treated equally. To address this, we design the robust personalization objective that adaptively shifts attention toward hard-to-learn users, improving overall robustness across the user distribution.

\vspace{3pt}
\noindent$\bullet$ \textbf{User-level filtering.}  
We identify hard-to-learn users as those with larger query losses. A high query loss indicates that even after adaptation, the model still performs poorly on that user, reflecting greater difficulty in capturing the preference. To emphasize these users, we introduce a filtering strategy. Given user query losses $\{\mathcal{L}_{\mathcal{T}_i}\}_i$, we set a threshold $\tau$ as the $(1-\rho)$ quantile, where $\rho \in (0,1]$ specifies the fraction of hardest users to retain. The objective is

\begin{equation}
\mathcal{A}_\rho(\{\mathcal{L}_{\mathcal{T}_i}\}_i) \;=\; \sum_{i=1}^n \tilde{\mathcal{L}}_{\mathcal{T}_i}, \quad 
\tilde{\mathcal{L}}_{\mathcal{T}_i} =
\begin{cases}
\mathcal{L}_{\mathcal{T}_i}, & \mathcal{L}_{\mathcal{T}_i} > \tau, \\
0, & \mathcal{L}_{\mathcal{T}_i} \le \tau .
\end{cases}
\end{equation}
Here, only the hardest $\rho n$ users contribute to the outer loop meta update, ensuring that optimization is guided by cases where the model struggles most, rather than by easy users it already fits well.

\vspace{3pt}
\noindent$\bullet$ \textbf{Soft reweighting with smoothing.}
However, only keeping the hardest users can destabilize training and overemphasize the most difficult users at the cost of overall performance. To mitigate this, we introduce a soft reweighting scheme that smoothly decreases the weight of users below the threshold, while keeping a higher weight for those above it. Formally, each task loss is reweighted as

\begin{equation}
\tilde{\mathcal{L}}_{\mathcal{T}_i} =\sigma\!\left(\tfrac{\mathcal{L}_{\mathcal{T}_i}-\tau}{\gamma}\right) \, \mathcal{L}_{\mathcal{T}_i}, 
\end{equation}
where $\sigma(\cdot)$ is the sigmoid function and smoothing parameter $\gamma>0$ controls the sharpness of the transition. The resulting objective is:

\begin{equation}
\mathcal{A}_{\rho,\gamma}(\{\mathcal{L}_{\mathcal{T}_i}\}_i) = \sum_{i=1}^n \sigma\!\left(\tfrac{\mathcal{L}_{\mathcal{T}_i}-\tau}{\gamma}\right) \, \mathcal{L}_{\mathcal{T}_i}.
\end{equation}
This soft reweighting preserves emphasis on hard-to-learn users while preventing easy ones from being completely discarded, yielding smoother gradients and more stable training.

\section{Experiments}
We conduct experiments to answer five research questions:

\begin{itemize}[itemsep=0pt, topsep=0pt, parsep=0pt, partopsep=0pt, leftmargin=*]
\item \textbf{RQ1:} How does our proposed MRM compare in performance to existing personalized reward modeling methods?
\item \textbf{RQ2:} How does MRM perform in terms of robustness across users compared to previous methods?
\item \textbf{RQ3:} How do the components of MRM affect the performance?
\item \textbf{RQ4:} How does MRM adapt to unseen users under different numbers of few-shot examples compared to existing methods?
\item \textbf{RQ5:} How does MRM compare to existing methods in terms of efficiency and scalability with respect to the number of users?
\end{itemize}

\subsection{Experimental Settings}
We describe the experimental settings in this section, which covers the datasets, the baselines, the evaluation, and the implementation.

\subsubsection{\textbf{Datasets}}
Following prior work~\cite{bose2025LORE}, we evaluate on two datasets with user-level preference annotations: PRISM~\cite{kirk2024PRISM} and Reddit TLDR~\cite{2020Reddittldr}. PRISM contains multi-turn conversations with LLMs, where participants provide preference feedback over model responses. Each instance consists of a dialogue context paired with candidate responses, along with user judgments indicating which response better aligns with their preference. After filtering users with fewer than 6 dialogues, we obtain 1,287 users, each with an average of 6 dialogues. Reddit TLDR consists of posts, two candidate summaries, and a preference label, with annotator IDs serving as user identifiers. Removing users with fewer than 50 annotations leaves 40 users, each contributing about 3,750 labeled pairs.

To simulate realistic personalization, we split users into two equal groups: \textit{\textbf{seen}} and \textit{\textbf{unseen}}. Each user’s data is further equally divided into train and test sets. The seen-train split is used for training, while the unseen-train split is reserved for few-shot adaptation. To reflect limited feedback per user in Reddit TLDR, we restrict the seen-train split to 100 or 150 pairs per user and the unseen-train split to 50 pairs per user, while keeping full test sets. We denote these two settings as \textbf{Reddit TLDR (100 examples)} and \textbf{Reddit TLDR (150 examples)}. For \textbf{PRISM}, since each user has relatively few dialogues, we retain all available data for both train and test.

\subsubsection{\textbf{Baselines}}
We compare MRM against representative methods from both personalized input and personalized parameter approaches. Here, we briefly introduce each baseline. Following prior work~\cite{bose2025LORE}, most methods use the same embedding extracted from a pretrained reward model~\cite{liu2024skyworkrewardbagtricksreward}. 
1) \textbf{Skywork-Reward}~\cite{liu2024skyworkrewardbagtricksreward, liu2025skyworkrewardv2} refers to a series of open-source pretrained reward models (V1 and V2) based on Llama-3.1-8B-Instruct~\cite{grattafiori2024llama3herdmodels}. These models directly score prompt--response pairs.
2) \textbf{BT} trains a single model with Bradley–Terry loss~\cite{bradley1952BTloss}, serve as a non-personalized baseline.
3) \textbf{GPO}~\cite{zhao2024GPO} learns group preferences by training a transformer that predicts the current preference attending to embeddings of prior preference pairs.
4) \textbf{VPL}~\cite{poddar2024VPL} encodes past user preferences into a user-specific embedding, which conditions the reward model for personalization.
5) \textbf{PAL}~\cite{chen2025PAL} represents each user as a weight distribution over a finite set of preference prototypes, enabling the reward model to capture individual variation.
6) \textbf{LoRe}~\cite{bose2025LORE} applies a low-rank decomposition of reward functions, representing individual preferences as weighted combinations of basis reward functions, with user-specific weights learned separately.
7) \textbf{SynthesizeMe}~\cite{ryan-etal-2025-synthesizeme} builds user personas by reasoning over prior preferences and synthesizing prompts to guide LLM judgments. We evaluate both in-context learning (ICL) and fine-tuned (FT) variants based on the same backbone of Llama-3.1-8B-Instruct~\cite{grattafiori2024llama3herdmodels}. Since this method requires repeated generations and is slow on the full test set, we evaluate each user on a randomly sampled quarter of their test instances for efficiency.

\subsubsection{\textbf{Evaluation Settings}}
We evaluate methods using user-level accuracy on test response pairs. 
For a user $u_i$ with test set $\mathcal{D}^{test}_i = \{(x, y_c, y_r)\}$, accuracy is:

\begin{equation}
\text{Acc}(u_i) = \frac{1}{|\mathcal{D}^{test}_i|} 
\sum_{(x,y_c,y_r) \in \mathcal{D}^{test}_i} 
\mathbf{1}\big(R_\phi(x,y_c) > R_\phi(x,y_r)\big),
\end{equation}
where $R_\phi$ is the reward model and $\mathbf{1}(\cdot)$ equals 1 if the inequality holds and 0 otherwise.  
We report the average accuracy over all users $U$:

\begin{equation}
\text{Avg. Acc.} = \tfrac{1}{|U|} \sum_{i=1}^{|U|} \text{Acc}(u_i).
\end{equation}
Each experiment is repeated 20 times with different seen and unseen splits, and we report the mean and standard deviation.

\subsubsection{\textbf{Implementation Details}}
Consistent with the baselines, MRM is implemented based on embeddings from Skywork-Reward~\cite{liu2024skyworkrewardbagtricksreward, liu2025skyworkrewardv2}, with each user’s reward model represented as a weighted combination of 2 base reward functions. Experiments are run on NVIDIA RTX A5000 GPU, and the number of training epochs is selected based on held-out evaluation on each dataset. The inner loop performs one adaptation step per user, and meta-optimization uses a batch size of 2 users. Both loops are optimized with Adam~\cite{kingma2017Adam}. For PRISM, the meta and inner learning rates are set to 1e-3; for Reddit TLDR, 5e-3. For each user, 10\% of the training data is used as the support set and 90\% as the query set. In RPO, the proportion of hard-to-learn users $\rho=0.5$, and the smoothing parameter $\gamma=0.5$.

\begin{table*}[]
\centering
\caption{Performance comparison between our proposed MRM and baselines. The numbers report user-level accuracy (\%) averaged over 20 random splits, with mean and standard deviation. Bold numbers indicate the best performance in each column, while the second-best are underlined. * implies the improvements over the best baselines are statistically significant ($p$-value < 0.05). \% Improvement indicates the gain relative to the best-performing baseline. $^{\dagger}$ For SynthesizeMe, each user is evaluated on a randomly sampled quarter of their test set due to the high cost of in-context inference and retries.}
\setlength{\tabcolsep}{1.8mm}{
\resizebox{1\textwidth}{!}{
\begin{tabular}{cccclccclccc}
\hline
\multirow{2}{*}{\textbf{Method}} & \multicolumn{3}{c}{\textbf{PRISM}}                              &  & \multicolumn{3}{c}{\textbf{Reddit TLDR (100 examples)}}         &  & \multicolumn{3}{c}{\textbf{Reddit TLDR (150 examples)}}         \\ \cline{2-4} \cline{6-8} \cline{10-12} 
                                 & \textbf{Seen}       & \textbf{Unseen}     & \textbf{Overall}    &  & \textbf{Seen}       & \textbf{Unseen}     & \textbf{Overall}    &  & \textbf{Seen}       & \textbf{Unseen}     & \textbf{Overall}    \\ \hline
\textbf{Skywork-Reward V1}~\cite{liu2024skyworkrewardbagtricksreward}          & 59.5 ± 0.5          & 59.5 ± 0.4          & 59.5 ± 0.5          &  & 62.6 ± 0.8          & 62.6 ± 0.8          & 62.6 ± 0.8          &  & 62.6 ± 0.8          & 62.6 ± 0.8          & 62.6 ± 0.8          \\
\textbf{Skywork-Reward V2}~\cite{liu2025skyworkrewardv2}          & 60.5 ± 0.4          & 60.1 ± 0.5          & 60.3 ± 0.3          &  & 64.4 ± 0.7          & 64.5 ± 0.7         & 64.4 ± 0.4       &  &  64.4 ± 0.7         & 64.5 ± 0.7         & 64.4 ± 0.4        \\
\textbf{BT}~\cite{bradley1952BTloss}                      & 64.3 ± 0.5          & 64.2 ± 0.6          & { 64.4 ± 0.3}    &  & 67.8 ± 1.1          & 67.8 ± 0.8          & 67.8 ± 0.8          &  & 68.1 ± 1.1          & 68.2 ± 0.8          & 68.1 ± 0.8          \\
\textbf{GPO}~\cite{zhao2024GPO}                     & 64.2 ± 1.1          & 64.2 ± 1.0          & 64.2 ± 1.0          &  & 68.0 ± 1.1          & 68.0 ± 1.2          & 68.0 ± 0.5          &  & { 68.5 ± 1.1}    & 68.6 ± 1.2          & { 68.6 ± 0.6}    \\
\textbf{VPL}~\cite{poddar2024VPL}                     & { 64.6 ± 1.0}    & 64.0 ± 1.2          & 64.3 ± 1.1          &  & 67.6 ± 1.1          & 67.7 ± 1.0          & 67.6 ± 1.0          &  & 67.7 ± 1.2          & 67.9 ± 1.0          & 67.8 ± 1.0          \\
\textbf{PAL}~\cite{chen2025PAL}                     & 61.7 ± 1.0          & 61.4 ± 0.8          & 61.5 ± 0.8          &  & 65.8 ± 1.0          & 65.9 ± 1.1          & 64.5 ± 0.7          &  & 66.3 ± 1.2          & 66.7 ± 0.9          & 66.5 ± 0.7          \\
\textbf{LoRe}~\cite{bose2025LORE}                    & 63.0 ± 0.9          & 63.1 ± 0.8          & 63.0 ± 0.8          &  & 68.1 ± 1.1          & { 68.6 ± 1.2}    & { 68.3 ± 0.5}    &  & { 68.5 ± 1.0}    & { 68.8 ± 1.1}    & { 68.6 ± 1.0}    \\
\textbf{SynthesizeMe (ICL)}$^{\dagger}$~\cite{ryan-etal-2025-synthesizeme}      & 63.9 ± 1.2          & 63.4 ± 1.2          & 63.7 ± 1.2          &  & 66.6 ± 2.2          & 66.3 ± 2.2          & 66.5 ± 2.0          &  & 66.7 ± 2.2          & 66.3 ± 2.3          & 66.5 ± 2.3          \\
\textbf{SynthesizeMe (FT)}$^{\dagger}$~\cite{ryan-etal-2025-synthesizeme}       & 64.3 ± 1.0          & { 64.5 ± 1.1}    & { 64.4 ± 1.0}    &  & { 68.2  ± 1.3}   & 68.0 ± 1.3          & 68.1 ± 1.3          &  & 68.2  ± 1.3         & 68.0 ± 1.3          & 68.1 ± 1.3          \\
\textbf{MRM (Skywork-Reward V1)}                     & {\ul 64.8 ± 0.4} & { \ul 64.9 ± 0.4} & {\ul 64.9 ± 0.2} &  & {\ul 68.7 ± 1.1} & {\ul 69.0 ± 0.8} & {\ul 68.8 ± 0.4} &  & {\ul 69.0 ± 1.1} & {\ul 69.5 ± 0.8} & {\ul 69.3 ± 0.3} \\
\textbf{MRM (Skywork-Reward V2)}    & \textbf{65.3 ± 0.6*} & \textbf{65.2 ± 0.5*} & \textbf{65.3 ± 0.3*} &  & \textbf{69.6 ± 0.9*}  & \textbf{69.6 ± 0.8*} & \textbf{69.6 ± 0.3*} &  & \textbf{69.7 ± 0.8*}  & \textbf{69.8 ± 0.9*}  & \textbf{69.7 ± 0.3*}  \\
\textbf{\% Improvement}             & 1.1\% & 1.1\% & 1.4\% &  & 2.1\% & 1.5\% & 1.9\% &  & 1.8\% & 1.5\% & 1.6\% \\ \hline
\end{tabular}}}
\label{tab: overall}
\vspace{-1em}
\end{table*}

\subsection{Overall Performance (RQ1)}

We compare MRM with baseline methods on personalized preference learning. Results in Table~\ref{tab: overall} yield the following observations:

\noindent$\bullet$ Across all methods, the gap between seen and unseen users is small, and most personalized methods offer only limited improvements over BT. This suggests that existing approaches struggle to extract strong user-specific signals in the data-scarce setting. In contrast, MRM consistently outperforms the BT baseline. This suggests that MRM's meta-learning objective is uniquely capable of leveraging limited data for effective personalization

\noindent$\bullet$ On PRISM, personalized input methods (GPO, VPL, and SynthesizeMe) slightly outperform personalized parameter methods (PAL and LoRe). This pattern supports our earlier analysis regarding the limitations of personalized parameter approaches: learning user-specific parameters from scratch is brittle in few-shot scenarios, where sparse feedback inevitably leads to overfitting and poor generalization to unseen users. Meanwhile, even personalized input methods do not surpass the non-personalized BT baseline, consistent with prior findings~\cite{ryan-etal-2025-synthesizeme} that personalization fails when per-user feedback is scarce. 

\noindent$\bullet$ On Reddit TLDR, personalized input methods achieve clear gains over the BT model because the dataset provides richer per-user feedback. This suggests that personalized input methods benefit from richer feedback per user, which aligns with the limitations noted earlier. Among personalized parameter methods, LoRe also shows notable improvement and becomes more competitive, likely because the smaller number of users allows it to better balance shared structures with user-specific modules in training.

\noindent$\bullet$ Overall, MRM consistently achieves the strongest performance across datasets and settings, yielding relative improvements of around 1.5\% over the best-performing baselines. This advantage comes from two key factors: 1) The meta weight initialization enables efficient few-shot personalization and strong generalization to unseen users. Unlike personalized input methods that rely on abundant per-user data or parameter-based methods that struggle with large user populations, MRM adapts effectively from only a few examples (refer to \S~\ref{sec:fewshot} for empirical evidence). 2) By incorporating the RPO, MRM places greater emphasis on hard-to-learn users and thus achieves robust performance across diverse users (see \S~\ref{sec:robustness} for detailed analysis).

\begin{figure}[t]
\setlength{\abovecaptionskip}{-0.05cm}
\setlength{\belowcaptionskip}{-0.20cm}
\centering 
\includegraphics[width=1\linewidth]{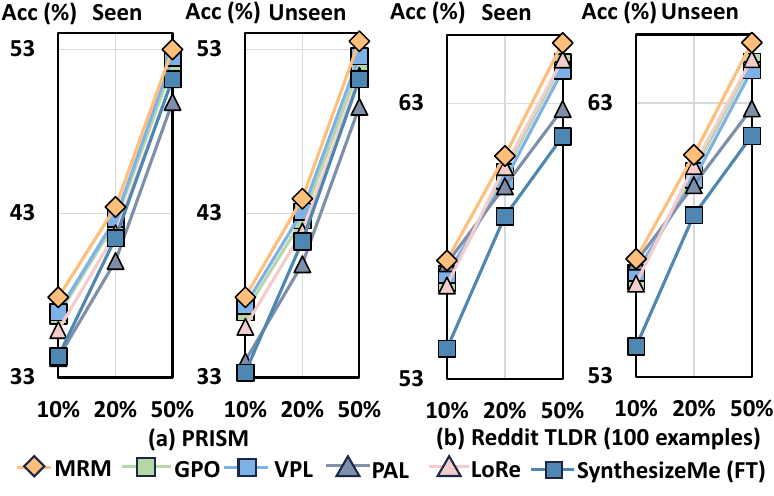}
\caption{Performance of average accuracy on the worst 10\%, 20\%, and 50\% of users for (a) PRISM and (b) Reddit TLDR with 100 examples. MRM consistently outperforms baselines on all proportions of worst users, showing stronger robustness.}
\label{Fig: hard}
\end{figure}

\subsection{In-depth Analysis}
The overall results show that MRM achieves strong performance; however, they do not fully reveal the source of the gains. Therefore, we further analyze the robustness of MRM, the contribution of core components, and few-shot adaptation to unseen users.

\subsubsection{\textbf{User Robustness (RQ2)}}
\label{sec:robustness}
Most prior methods optimize average performance across users, whereas true personalization demands consistent performance for every individual. To enhance robustness across users, we introduce the RPO to explicitly emphasize hard-to-learn users during meta optimization.

We evaluate robustness by measuring the average accuracy of the worst 10\%, 20\%, and 50\% of users on PRISM and Reddit TLDR (100 examples). Personalized input methods are shown as squares, and Personalized parameter methods as triangles. The results in Figure~\ref{Fig: hard} show that: 1) Personalized input methods and parameter-based methods both show sharp drops on the hardest users, with PRISM results even falling below random choice. SynthesizeMe (FT), relying on in-context learning, is more unstable and performs the worst. These results indicate that existing methods generally struggle to maintain performance for hard-to-learn users, as they are not explicitly designed to handle diverse or atypical preference patterns. 2) In contrast, MRM consistently outperforms all baselines across the hardest user subsets. This improvement arises from RPO, which dynamically reweights user losses to focus outer-loop updates on the most difficult users. By learning from these challenging cases, MRM avoids overfitting to easy users and achieves more robust performance across the entire user distribution. This explains why MRM adapts more reliably to diverse users and shows clear advantages in robustness, a key factor behind its overall gains.

\begin{figure}[t]
\setlength{\abovecaptionskip}{-0.05cm}
\setlength{\belowcaptionskip}{-0.20cm}
\centering
\includegraphics[width=1\linewidth]{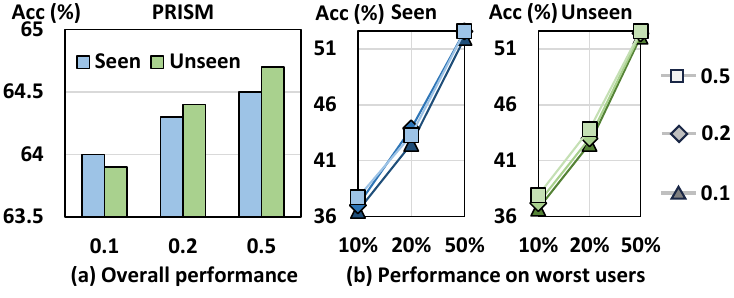}
\caption{Effect of threshold ratio on PRISM.
(a) Overall accuracy with different threshold ratios ($\rho=0.1, 0.2, 0.5$).
(b) Accuracy on the worst $k\%$ of users ($k=10,20,50$).}
\label{Fig: ratio}
\end{figure}

\begin{figure}[t]
\setlength{\abovecaptionskip}{-0.05cm}
\setlength{\belowcaptionskip}{-0.10cm}
\centering
\includegraphics[width=1\linewidth]{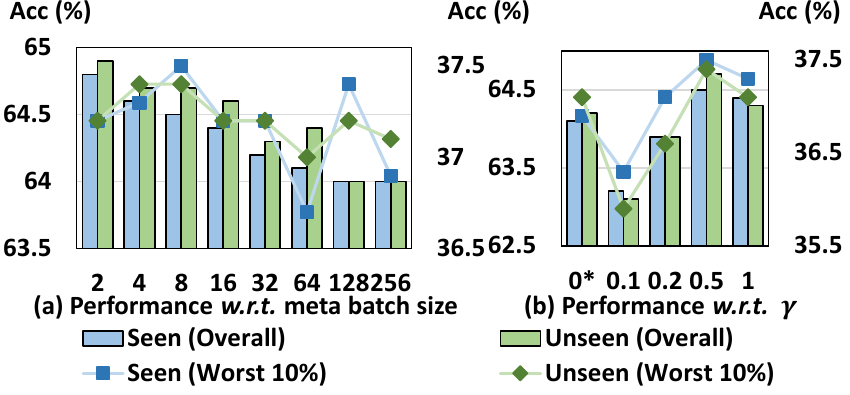}
\caption{Performance with respect to (a) meta batch size and (b) smoothing parameter $\gamma$. In (b), $\gamma{=}0^*$ denotes hard filtering.}
\label{Fig: batchgamma}
\end{figure}

\begin{figure*}[t]
\setlength{\abovecaptionskip}{-0.05cm}
\centering
\includegraphics[width=1\linewidth]{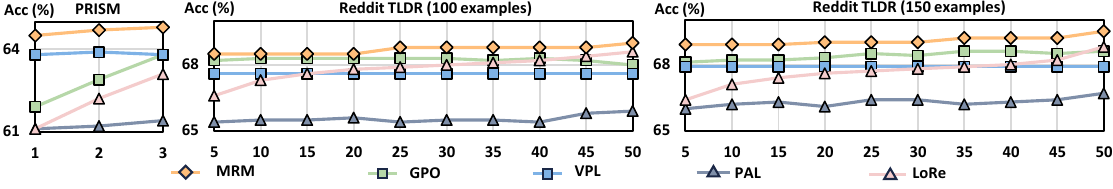}
\caption{Performance of few-shot adaptation on unseen users. We vary the number of few-shot examples for each user (x-axis) and report accuracy on three settings. MRM consistently outperforms baselines and shows stronger gains with more examples.}
\vspace{-1em}
\label{Fig: fewshot}
\end{figure*}

\begin{table*}[]
\setlength{\belowcaptionskip}{-0.20cm}
\centering
\caption{Ablation study on key components of MRM.}
\setlength{\tabcolsep}{1mm}{
\resizebox{1\textwidth}{!}{
\begin{tabular}{lcccclcccc}
\hline
\multicolumn{1}{c}{\multirow{2}{*}{\textbf{Variants}}} & \multicolumn{4}{c}{\textbf{Seen}}                                                  &  & \multicolumn{4}{c}{\textbf{Unseen}}                                                \\ \cline{2-5} \cline{7-10} 
\multicolumn{1}{c}{}                                   & \textbf{Overall} & \textbf{Worst 10\%} & \textbf{Worst 20\%} & \textbf{Worst 50\%} &  & \textbf{Overall} & \textbf{Worst 10\%} & \textbf{Worst 20\%} & \textbf{Worst 50\%} \\ \hline
\textbf{(0): MRM}                                      & 64.8 ± 0.4       & 37.9 ± 1.3          & 43.4 ± 0.9          & 52.8 ± 0.6          &  & 64.9 ± 0.4       & 37.9 ± 0.8          & 43.9 ± 0.7          & 53.2 ± 0.7          \\
\textbf{(1): MRM \textit{w/o} meta-learning formulation}        & 63.1 ± 0.9       & 35.9 ± 1.1          & 41.8  ± 0.8         & 51.2  ± 0.6         &  & 63.1 ± 0.8       & 36.1 ± 0.9          & 41.9 ± 0.8          & 51.4 ± 0.7          \\
\textbf{(2): MRM \textit{w/o} RPO}                              & 64.5 ± 0.5       & 37.5 ± 1.3          & 43.0 ± 0.9          & 52.5 ± 0.6          &  & 64.5 ± 0.6       & 37.2 ± 0.9          & 43.5 ± 0.7          & 52.7 ± 0.7          \\
\textbf{(3): MRM \textit{w/o} basis combination}                & 63.8 ± 0.4       & 37.1 ± 1.4          & 42.8 ± 1.0          & 52.3 ± 0.6          &  & 63.8 ± 0.6       & 37.2 ± 0.9          & 42.8 ± 0.7          & 52.5 ± 0.6          \\ \hline
\end{tabular}}}
\label{tab: ablation}
\vspace{-1em}
\end{table*}

\subsubsection{\textbf{Analysis of RPO (RQ2)}}
We analyze the impact of the robust personalization objective by examining three parameters: (1) \uline{the threshold ratio}, specifying the fraction of the hard users in each batch; (2) \uline{the meta batch size}, controlling the number of users per update; and (3) \uline{the smoothing parameter}, regulating the sharpness of reweighting. All experiments are based on PRISM.

\vspace{3pt}
\noindent$\bullet$ \textbf{Effect of threshold ratio.}  
We vary the threshold ratio across three representative values of 0.1, 0.2, and 0.5, with a meta batch size of 8, and evaluate both overall accuracy and the accuracy of the worst 10\%, 20\%, and 50\% of users. The results in Figure~\ref{Fig: ratio} reveal several key observations. 1) On overall performance, increasing the ratio from 0.1 to 0.5 yields a monotonic improvement. This suggests that setting the threshold too small harms performance, likely because excessive emphasis on the hardest users shifts the optimization focus, weakening learning on other users that are also important. 2) On worst users, ratios of 0.2 and 0.5 perform comparably and outperform 0.1, while at 50\% all settings converge. This shows that overemphasizing the hardest users does not actually improve them, but instead reduces stability and overall effectiveness. In summary, a ratio of 0.5 provides the best trade-off between focusing on the hardest users and maintaining generalization, resulting in both robustness and overall stability.

\vspace{3pt}
\noindent$\bullet$ \textbf{Effect of meta batch size.}  
We vary the number of users sampled per outer loop update and evaluate both overall accuracy and accuracy on the worst 10\% of users. As shown in Figure~\ref{Fig: batchgamma} (a), we have the following observations. 1) Overall accuracy decreases steadily as batch size increases, indicating that optimizing over many users simultaneously becomes harder, and the reweighting in RPO further reduces training stability, leading to weaker performance. 2) For the worst 10\%, accuracy improves as batch size grows from 2 to 8 but drops beyond that point. This suggests that a moderate batch size helps RPO identify and emphasize harder users, since larger batches provide a better approximation of globally hard users. However, overly large batches complicate optimization and reduce effectiveness. Overall, these results highlight that relatively small batch sizes offer a better trade-off between stability and robustness.

\vspace{3pt}
\noindent$\bullet$ \textbf{Effect of smoothing parameter.}  
We compare hard filtering and soft reweighting under different smoothing parameters $\gamma$ with a meta batch size of 8 and evaluate both overall accuracy and accuracy on the worst 10\% of users. We have the following observations from Figure~\ref{Fig: batchgamma} (b). 1) When $\gamma$ is very small, performance is poor for both overall and worst users; as $\gamma$ increases, accuracy improves and peaks around 0.5, then declines slightly at 1. This shows that too small a $\gamma$ destabilizes training by placing nearly all weight on the hardest users, while too large a $\gamma$ over-smooths the weighting and weakens the intended emphasis. 2) Hard filtering performs better than near-zero $\gamma$ but still worse than $\gamma{=}0.5$. A plausible explanation is that near-zero $\gamma$ does not replicate hard filtering: instead of discarding easy users, it rescales their losses, altering gradient magnitudes and causing instability. By contrast, hard filtering drops users below the threshold without changing the scale of losses. Overall, smoothing of 0.5 achieves the best balance: it avoids the instability of sharp weighting and the over-smoothing caused by large $\gamma$, while also outperforming pure hard filtering.

\subsubsection{\textbf{Ablation Study (RQ3)}}
To thoroughly investigate the contribution of each component in MRM, we conduct an ablation study by removing each component individually. We evaluate the following variants: (0)\uline{``MRM''}: our proposed method, (1) \uline{``MRM w/o meta-learning formulation''}, which independently trains weights for each user without meta-learning; (2) \uline{``MRM w/o RPO''}, which removes the robust personalization objective in outer loop optimization; and (3) \uline{``MRM w/o basis combination''}, which replaces the model with a single MLP updated in both inner and outer loops. We evaluate all variants on both datasets and report the accuracies of overall and worst users in Table~\ref{tab: ablation}.
We have the following observations. 1) Removing the meta-learning formulation causes the largest drop, showing that meta-learned initialization is key for few-shot personalization and adapting to unseen users. 2) Removing RPO slightly lowers accuracy, indicating that handling hard-to-learn users improves robustness, which in turn benefits not only the worst-case users but also the overall performance. 3) Removing basis combination also reduces performance, suggesting that decomposing the reward model into multiple basis reward functions does help to capture diverse preferences better.

\vspace{3pt}
\noindent$\bullet$ \textbf{Effect of number of base reward functions.}
We further analyze the impact of the number of base reward functions $K$ by varying $K$ over a wide range. We have the following observations. 
1) Increasing $K$ from 1 to a small value (e.g., $K=2$) leads to a slight improvement, indicating that combining multiple base functions helps capture user preference variations. 
2) However, further increasing $K$ does not lead to noticeable performance gains, with results remaining largely stable across different values of $K$. 
3) These findings suggest that MRM is not sensitive to the choice of $K$ once $K>1$, and a small number of base functions is sufficient for effective personalization. This supports our use of a minimal setting $K=2$ for efficiency.

\subsubsection{\textbf{Few-shot Adaptation (RQ4)}}
\label{sec:fewshot}
This experiment evaluates how different methods adapt to unseen users when provided with different numbers of few-shot examples. For each unseen user, we vary the number of few-shot examples and report accuracy on their test data. Both SynthesizeMe variants are not included here because inserting all examples exceeds the input length limit.

The results in Figure~\ref{Fig: fewshot} lead to the following observations: 1) Accuracy improves steadily for all methods as the number of examples increases, showing that additional feedback consistently aids adaptation to unseen users. 2) Personalized input methods, such as GPO and VPL, achieve stronger accuracy across different numbers of examples. In contrast, personalized parameter methods perform worse in data-scarce settings, with LoRe showing noticeable gains only when more examples are available. This pattern suggests that personalized input methods can leverage additional examples more effectively from the input side, while personalized parameter approaches must learn new user-specific structures, which is harder with limited data. 3) MRM consistently outperforms baselines, especially with very limited feedback. This improvement arises because the meta-learned weight initialization extracts shared patterns across seen users, providing a strong starting point that enables rapid adaptation to unseen users. As more feedback becomes available, MRM continues to refine user-specific parameters, leading to steady performance gains. This confirms MRM’s strength in few-shot personalization and aligns with the overall findings.

\begin{figure}[t]
\setlength{\abovecaptionskip}{-0.05cm}
\setlength{\belowcaptionskip}{-0.20cm}
\centering 
\includegraphics[width=1\linewidth]{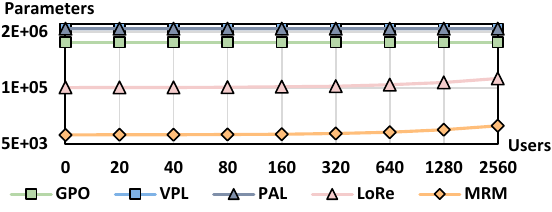}
\caption{Number of trainable parameters as the number of users increases for different methods.}
\label{Fig: param}
\end{figure}

\subsubsection{\textbf{Efficiency and Scalability (RQ5)}}
To evaluate the efficiency and scalability of different methods, we analyze both parameter growth and computational cost with respect to user scale. We first examine how the number of trainable parameters changes as more users are introduced. For each method, we vary the number of users and report the total number of trainable parameters required. All methods are implemented following their original descriptions. For MRM, the parameter count includes both the shared base functions and the adapted weights stored for each user. 

The results in Figure~\ref{Fig: param} lead to the following observations: 1) Personalized parameter methods, including PAL and LoRe, show parameter growth that scales linearly with the number of users, reflecting the need to allocate user-specific modules for each individual. 2) Personalized input methods, such as GPO and VPL, maintain a constant number of parameters regardless of user scale, since user preferences are encoded through extra input rather than separate structures. 3) MRM consistently requires the smallest number of trainable parameters across all user settings. Although its parameter count increases as the user grows, the growth remains limited due to lightweight shared base reward functions and user-specific weights, making MRM the most parameter-efficient and scalable among all methods.

In terms of computational efficiency, personalized reward modeling inevitably involves user-specific learning. Compared to personalized parameter methods that train and maintain separate user-specific modules, MRM concentrates computation on shared base reward functions and a shared weight initialization. Although MRM adopts an MAML-style training procedure, the additional computation is limited, as meta-learning is applied only to the initialization of lightweight user-specific weights. At inference time, adaptation from the shared initialization is performed once per user, updating a small set of weights that can be stored and reused.

\section{Discussion}
We further discuss the limitations of MRM and broader considerations for deploying personalized reward modeling in real-world settings.

\subsection{Limitations}

Despite the strong performance of MRM, several limitations remain.

\vspace{3pt}
\noindent$\bullet$ \textbf{Static user preferences.}
Our current formulation assumes that user preferences remain static during adaptation. However, in real-world scenarios, user preferences may evolve over time due to changing contexts or long-term shifts in user intent. Extending the alignment method to handle such temporal dynamics would require not only continual or online adaptation mechanisms, but also appropriate datasets that capture time-varying user feedback. Since existing benchmarks mainly provide static snapshots of user preferences, modeling dynamic preference evolution remains an open challenge.

\vspace{3pt}
\noindent$\bullet$ \textbf{Computational complexity.}
MRM relies on a bi-level meta-learning framework, which introduces a higher computational cost compared to standard single-level approaches. However, this cost is primarily incurred during meta-training and is amortized at inference time. MRM is designed to mitigate this cost by sharing base reward functions across users and restricting personalization to a small set of lightweight user-specific weights. This design keeps both parameter and adaptation costs low in practice. Nevertheless, further improving the efficiency of meta-training remains an important direction.

\vspace{3pt}
\noindent$\bullet$ \textbf{Evaluation scope.}
We evaluate performance at the reward modeling level using preference prediction accuracy, which is a standard proxy in this setting. However, our experiments do not extend to downstream policy optimization or text generation. As a result, this metric does not directly reflect the quality or usefulness of generated responses in real-world applications and may not fully capture user satisfaction. This gap may limit the ability to fully assess the practical impact of personalization in real-world applications. Future work may integrate MRM with policy optimization methods and include human evaluation or downstream task-based metrics.

\subsection{Ethical and Privacy Considerations}

Personalized reward modeling offers finer-grained alignment with individual preferences, but it also introduces ethical and privacy challenges that must be carefully addressed.

\vspace{3pt}
\noindent$\bullet$ \textbf{Privacy.} 
To align with individual users, models must learn from specific feedback, which inevitably reflects personal values and habits. Even if we remove explicit user IDs, the model might still memorize or reveal sensitive private information through its responses. Therefore, strict data protection measures and clear user consent are essential for responsible deployment. Potential mitigation strategies include data anonymization, minimizing stored user-specific information, and limiting the retention of personalized signals during training.

\vspace{3pt}
\noindent$\bullet$ \textbf{Bias.} 
Optimizing for individual preferences risks reinforcing user biases or narrowing exposure to alternative perspectives. A reward model trained solely on a user’s past feedback may overfit to prior judgments and amplify preference extremes. Designing personalization mechanisms that balance individual alignment with diversity and exploration is an important ethical consideration. Possible approaches include incorporating regularization across users or introducing constraints that encourage preference diversity during training.

\vspace{3pt}
\noindent$\bullet$ \textbf{Safety.} 
While personalized alignment represents a fine-grained level of alignment, it should remain consistent with broader societal norms and safety constraints. This introduces a fundamental tension between individual preferences and shared ethical boundaries. In the current MRM framework, such constraints are not explicitly enforced. A possible direction is to incorporate global safety constraints or regularization terms during meta-training, ensuring that personalized reward functions remain within acceptable boundaries. Another approach is to combine personalized rewards with a shared safety model that filters or constrains unsafe outputs. Developing principled mechanisms to balance individual alignment with societal norms remains an important direction for future work.

\vspace{3pt}
\noindent$\bullet$ \textbf{Reward hacking.} 
Reward models are inherently susceptible to reward hacking, where the model exploits imperfections in the learned reward function to maximize scores without faithfully capturing user intent. This issue can be amplified in personalized settings, as reward functions are learned from limited and potentially narrow user feedback, making them more prone to mis-specification and overfitting. As a result, the model may learn shortcuts that align with the reward signal but not with the user's true preferences. Possible mitigation strategies include introducing regularization to prevent overfitting to narrow signals, incorporating diverse feedback sources, or applying external validation or safety checks to ensure alignment with intended behaviors.

\section{Conclusion and Future Work}
In this work, we addressed the critical limitations of existing personalized reward modeling paradigms, which fail to simultaneously achieve efficient adaptation to unseen users while maintaining fine-grained personalization under limited user feedback. We argue that overcoming these challenges necessitates a fundamental shift from fitting static user models to ``learning to learn'' adaptation. Guided by this perspective, we proposed Meta Reward Modeling. By optimizing a highly adaptable initialization, MRM enables the model to rapidly adapt to unseen users using only a few demonstrations. For each user, preference data is split into a support set for inner loop personalization and a query set for outer loop meta optimization of both the weight initialization and base functions. Furthermore, to ensure robustness under diverse user preferences, we introduced the Robust Personalization Objective, which dynamically emphasizes hard-to-learn users during meta-optimization. Extensive experiments demonstrate that our framework effectively solves the few-shot personalization challenge, offering a robust and scalable path for aligning LLMs with diverse human values.

Building on this work, several promising directions remain open for future exploration. 1) Extending MRM from reward modeling to direct policy optimization (e.g., via DPO) would enable end-to-end personalized generation and streamline the alignment pipeline. 2) Modeling evolving user preferences through continual or online adaptation could further improve personalization in dynamic environments. 3) Instead of passively relying on sparse history, models could actively query users to resolve ambiguity, thereby reducing the sample complexity required for adaptation. 4) Beyond explicit pairwise preferences, incorporating implicit or noisy behavioral signals may provide richer supervision for personalization. 5) Extending MRM to more expressive model parameterizations, such as full-model meta-learning, may further improve personalization capacity, provided that scalability and training stability can be maintained.

\section*{Acknowledgments}
The work described in this paper was supported by the Research Grants Council of Hong Kong (PolyU/15213323, PolyU/15207122, PolyU/15209724, PolyU/15205325) and PolyU internal grants (BDWP).


\bibliographystyle{unsrtnat} 
\bibliography{ref}


\appendix

\end{document}